\documentclass[sigconf]{acmart}

\usepackage{booktabs} 

\setcopyright{none}

\begin{document}
\title{Video Imagination from a Single Image with Transformation Generation}

\author{Baoyang Chen, Wenmin Wang, Jinzhuo Wang, Xiongtao Chen}
\affiliation{\institution{School of Electronics and Computer Engineering, Peking University}}

\begin{abstract}
In this work, we focus on a challenging task: synthesizing multiple imaginary videos given a single image. Major problems come from high dimensionality of pixel space and the ambiguity of potential motions. To overcome those problems, we propose a new framework that produce imaginary videos by transformation generation. The generated transformations are applied to the original image in a novel volumetric merge network to reconstruct frames in imaginary video. Through sampling different latent variables, our method can output different imaginary video samples. The framework is trained in an adversarial way with unsupervised learning. For evaluation, we propose a new assessment metric $RIQA$. In experiments, we test on 3 datasets varying from synthetic data to natural scene. Our framework achieves promising performance in image quality assessment. The visual inspection indicates that it can successfully generate diverse five-frame videos in acceptable perceptual quality.

\end{abstract}
%


\maketitle

\newcommand{\tabincell}[2]{\begin{tabular}{@{}#1@{}}#2\end{tabular}}
\section{Introduction}
\begin{figure}
	\includegraphics[width=3in]{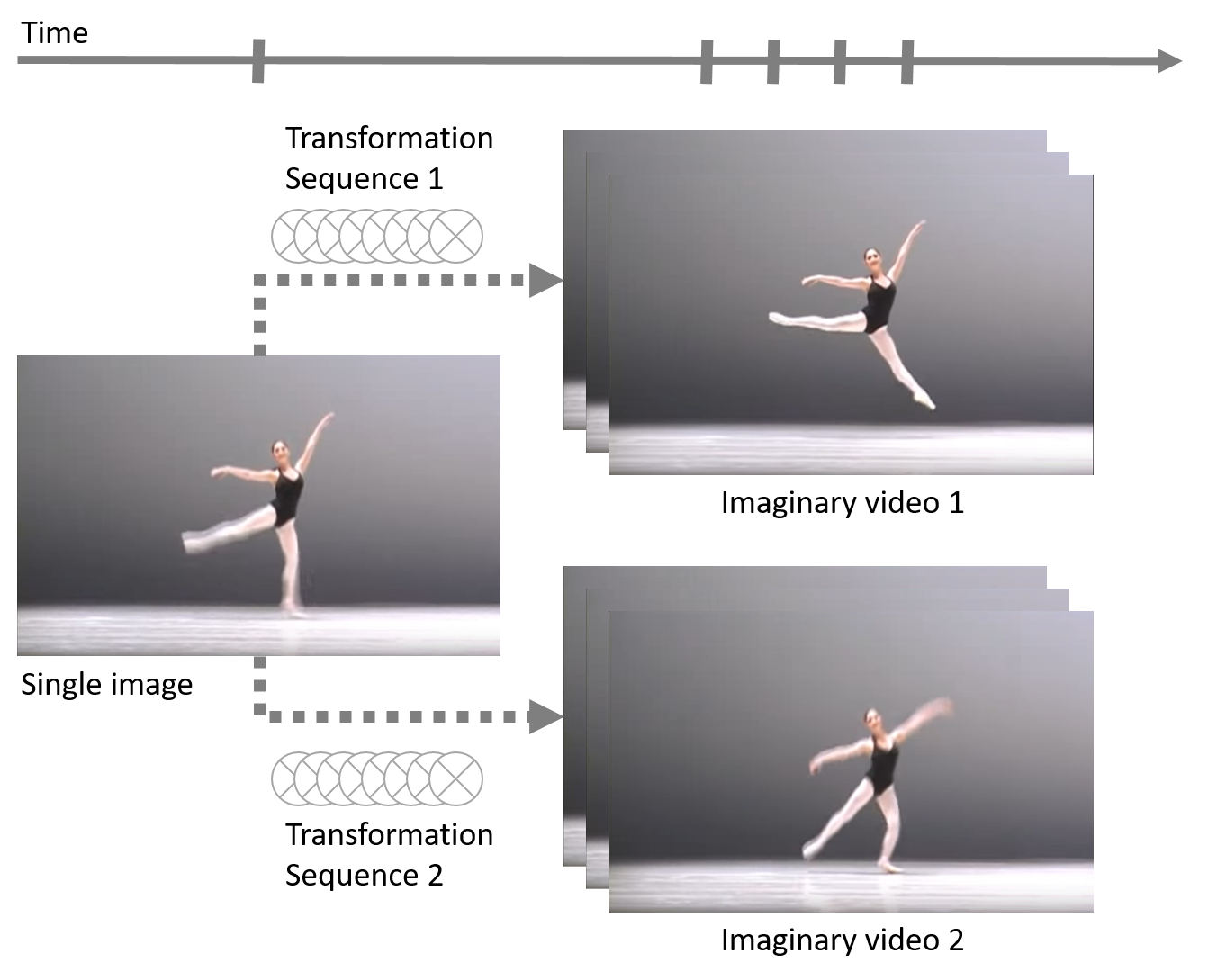}
	\caption{Synthesizing multiple imaginary videos from one single image. \textrm{For instance, given an image of a dancing ballerina, the videos of the dancer jumping higher or landing softly are both plausible imaginary videos. Those videos can be synthesized through applying a sequence of transformations to the original image}.}
	\label{fig:dance}
\end{figure}

Given an static image, humans can think of various scenes of what will happen next using their imagination.  For example, considering the ballerina in Figure \ref{fig:dance}, one can easily picture the scene of the dancer jumping higher or landing softly. In this work, we clarify the task as intimating human capability of \textit{\textbf{Video Imagination}}: 
synthesizing imaginary videos from single static image. 
This requires synthesized videos to be diverse and plausible. 
Although this study is still in its infancy, we believe video prediction and image reconstruction area can draw inspiration from it.

Compared to related tasks, e.g. video anticipation and prediction, there are more challenges for video imagination . Video imagination means to produce real high-dimension pixel values unlike low-dimension vectors in semantic anticipation. In addition, videos that are not identity to each other can all be reasonable, like imaginary video 1 and imaginary video 2 in Figure \ref{fig:dance}. So there is no precise ground truth as in common video prediction task. This intrinsic ambiguity makes regular criterion like MSE fails in evaluating whether the synthesized video is plausible. Moreover, compared to image generation, video synthesis needs to additionally model the temporal dependency that makes consecutive frames seem realistic. 

Pioneers make attempts. Dense trajectory \cite{walker2016uncertain} and optical flow \cite{pintea2014deja} have been used to model scene dynamics. Variational auto-encoder \cite{walker2016uncertain} and stochastic Markov-chain \cite{ranzato2014video} have been introduced to form generative model. However, those models still struggle in high dimension space where manifold is hardly tractable, and is unsatisfying in terms of criterion 

In this work, we present an end-to-end unsupervised framework with transformation generation for video imagination . 
Our key intuition is that we can model in transformation space instead of pixel space.
Since scenes in frames are usually consistent, we assume that the major motions between frames can be modeled by transformations.
If we reconstruct frame based on the original image and corresponding transformations, both scene dynamic and invariant appearance can be preserved well. 
In addition, we draw inspiration from image generation works \cite{radford2015unsupervised} that use adversarial training. We believe an elaborate critic network that understands both spatial and temporal dependency would serve as reasonable criterion. 

Based on the intuition and inspiration above, we design our framework focusing on model distributions in transformation space implicitly, and train it in adversarial way. 
In this framework, we generate transformation conditioned on the given image. Then we reconstruct each frame by applying the generated transformation to the given image. Latent variable is also introduced to enable diverse sampling. Casting this into an adversarial architecture,  we train our framework in a fully end-to-end fashion.

We believe this framework is a promising way to overcome existing challenges. As we build generation model in transformation space, it is more tractable to implicitly model the distribution of transformation. Conditioned on image makes generated transformation reasonable. The procedure of applying transformation to original image is similar to the insight of highway connection \cite{he2016identity}, and this helps the synthesized video maintaining sharp and clear. 
Also, the latent variable 
enables diverse imagination through sampling different transformations corresponding to different imaginary videos. 
Furthermore, there is nearly infinite resource for this unsupervised training. No label is needed, so every video clip can serve as a training sample. 


For evaluation, since there is no general evaluation metrics for this task, we employ image quality assessment method to evaluate the quality of reconstructed frames and present a relative image quality assessment ($RIQA$) to eliminate the scene difference. In experiments, we evaluate our idea on three datasets, including two artificial video datasets with simple motions and one natural scene video dataset with complex motions. 
The synthesized 4-frames video results show that our framework can produce diverse sharp videos with plausible motions. 
We compare our framework with some related methods and two custom baselines. The quantitative evaluation results suggest that our framework outperforms others including those methods that are given more prior information, and the qualitative comparison also shows the advance of our synthesized videos.

The primary contribution of this paper is developing a new end-to-end unsupervised framework to synthesize imaginary videos from single image. We also make brave attempt on new evaluation method. In section 2, we review related work. In section 3, we present our \textit{Video Imagination} video synthesis framework in details. In section 4, we illustrate new evaluation method $RIQA$ and show experiments and comparison.


%
%
\begin{figure*}[h]
	\includegraphics[width=7in]{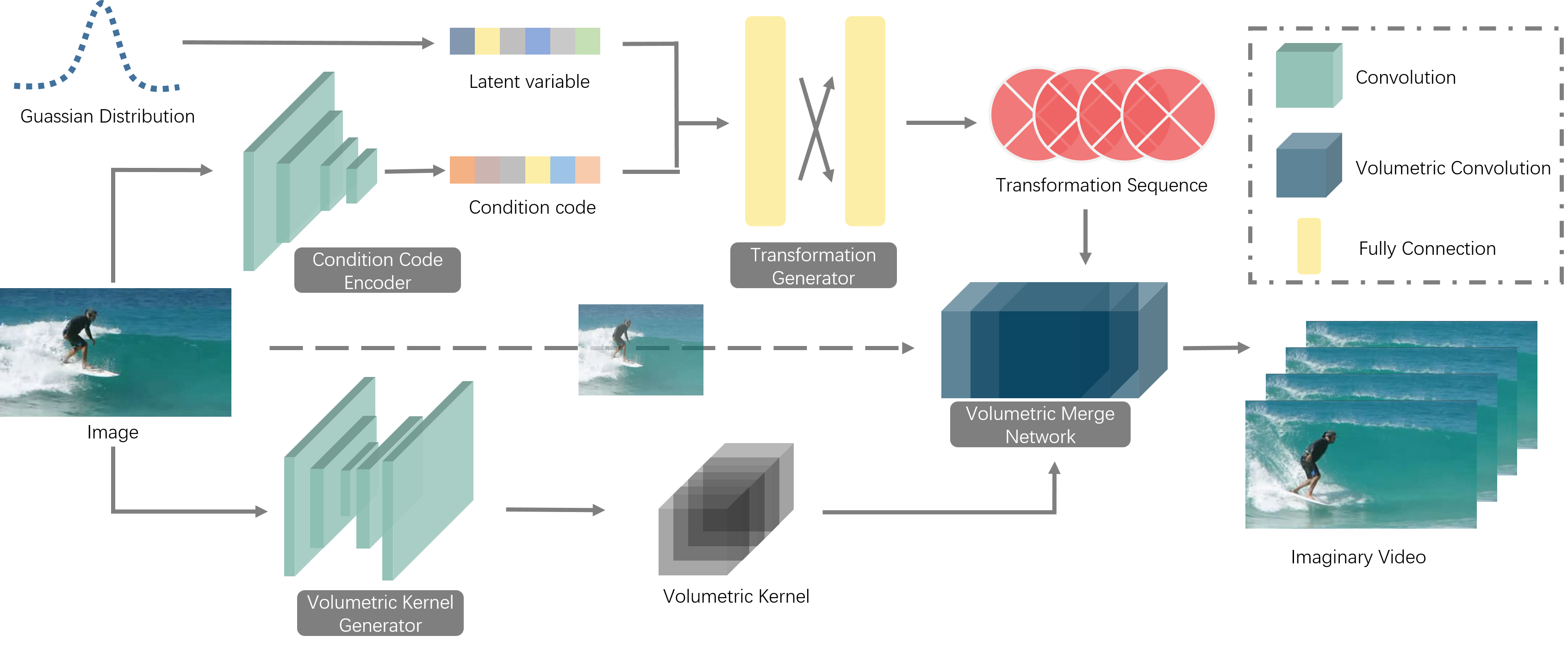}
	\caption{Pipeline of video imagination from single image. \textrm{In our framework, to produce one imaginary video, the input image is first encoded into a condition code and sent to transformation generator together with a latent variable. The generated transformation sequence is applied to input image later in volumetric merge network where frames are reconstructed with transformed images and volumetric kernels. Those four frames form one imaginary video. By sampling different latent variable from guassian distribution, our framework can produce diverse imaginary videos.} }
	\label{fig:model}
\end{figure*}

\section{Related work}

Although the works of future video synthesis from single image are rather little, our task shares common techniques with two related tasks: video prediction \cite{sutskever2014sequence} and image reconstruction \cite{vincent2008extracting}, where researchers have made impressive progress. In the following, we regard them as a universal visual prediction task, and review related works from different perspectives of approaches.

\textbf{Reconstruction in pixel space.} Early works of visual prediction focus on modeling and estimation in pixel space \cite{wang2004image} \cite{tan2006intra} \cite{wang2009mean}. These methods reconstruct images by calculating pixel values directly. With recent resurgence of deep networks, researchers tend to replace standard machine learning models with deep networks. In particular, \cite{kalchbrenner2016video} proposes a video pixel network and estimates the discrete joint distribution of the raw pixel values. \cite{srivastava2015unsupervised} uses LSTM network to learn representations of video and predict future frames from it. \cite{vondrick2016generating} employs adversarial training and generates video from scratch with deconvolution method \cite{zeiler2010deconvolutional}. 
A key issue in pixel-level prediction is the criterion metrics. A recent work \cite{mathieu2015deep} argues that standard mean squared error (MSE) criterion may fail with the inherently blurry predictions. They replace MSE in pixel space with a MSE on image gradients, leveraging prior domain knowledge, and further improves using a multi-scale architecture with adversarial training. 

\textbf{Mid-level tracking and matching.} To overcome the challenge of high dimensionality and ambiguity in pixel space, the prediction framework of mid-level elements gradually becomes popular. \cite{mahajan2009moving} explores a variation on optical flow that computes paths in the source images and copies pixel gradients along them to the interpolated images.  \cite{walker2014patch} combines the effectiveness of mid-level visual elements with temporal modeling for video prediction.\cite{ranzato2014video} defines a recurrent network architecture inspired from language modeling, predicting the frames in a discrete space of patch clusters. The input in \cite{walker2016uncertain} is a single image just like us, where the authors predict the dense trajectory of pixels in a scene with conditional variational autoencoder.

\textbf{Existing pixels utilization.} A insightful idea of improving the quality of prediction image is to utilize existing pixels. \cite{liu2017video} synthesizes video frames by flowing pixel values from existing ones through voxel flow. \cite{xue2016visual} outputs the difference image, and produces the future frame by sum up the difference image and raw frame. \cite{de2016dynamic} and \cite{finn2016unsupervised} share a similar methods with us of applying filters to raw frames to predict new frames, and they provide the validation of gradients flow through filters.

\textbf{Generation model evolutions.} Traditional works treat visual prediction as a regression problem. They often formulate prediction tasks with machine learning techniques to optimize the correspondence estimation \cite{kitani2012activity,lan2014hierarchical,hoai2014max}. With the development of deep networks, community of visual prediction has begun to produce impressive results by training variants of neural network structures to produce novel images and videos \cite{xie2016synthesizing,gregor2015draw,xie2016deep3d,zhou2016view}. The probabilistic models become popular again. More recently, generative adversarial networks (GANs) \cite{goodfellow2014generative} and variational autoencoders \cite{kingma2013auto} have been used to model and sample from distributions of natural images and videos  \cite{radford2015unsupervised,denton2015deep,yan2016attribute2image}. Our proposed algorithm is based on GAN, but unlike previous works starting with a simple noise, we force our generation model conditioned on the given image, which benefits to generate reasonable transformation.

To the best of our knowledge there are no existing model that can produce multiple videos given one single image. Perhaps the most similar works to our task are \cite{xue2016visual,walker2016uncertain}, where both works aim to build a probabilistic model of future given an image. But \cite{xue2016visual} only outputs one frame and \cite{walker2016uncertain} just produce optical flows.

Also, note a concurrent work that learns to predict in transformation space is \cite{van2017transformation}, where the authors predict the new frames by predicting the following affine transformations. But their task is to generate frames from sequence of frames while ours is to synthesize imaginary videos given a single image. In addition, our work differs in that there methods are close to a regression problem as to predict precise future frames, but our task requires a probabilistic view and aims at generating multiple videos.

\section{Approach}

Rather than struggle in high-dimension pixel space, our idea is to model in transformation space for video imagination: to take one single image as input and synthesize imaginary videos that picture multiple plausible scene change of that image. Figure \ref{fig:model} shows the pipeline of output one imaginary video in our framework.

In our framework, firstly, we send latent variable and condition code encoded from image into \textbf{transformation generator}, which outputs a group of transformation sequences. Secondly, we apply those transformation sequences to the original image and reconstruct frames through a \textbf{volumetric merge network}. Finally, we combine frames as an imaginary video then use \textbf{video critic network} to achieve adversarial training. 
In the following subsections, we firstly give a problem description; then we describe the details of those crucial parts and its implementation. 

\subsection{Problem definition  }

Firstly we use formulations to describe this task: given an image $X$, outputs $m$ imaginary videos $\hat V$ 
corresponding to different reasonable motions. Each imaginary video contains $T$ consecutive frames $f_T$.

Ideally, we would like to model the distribution $P(V \mid X)$ of all possible imaginary $V$ given $X$. Practically, we aim to train a neural network $T_{\theta}$ with parameters $\theta$, which implicitly models a distribution $P_T(V \mid X)$. Through training, we expect $P_T(V \mid X)$ to converge to a good estimate of $P(V \mid X)$. $T_{\theta}(X)$ yields a sample $\hat V$ drawn from $P_T(V \mid X)$, so we have
\begin{equation}
\hat V = T_{\theta}(X) \sim P_T(V \mid X)
\end{equation}

Instead of directly modeling in pixel space, we choose to model distribution in transformation space. We build this model based on a key assumption that the major motions between frames can be modeled by transformations. That means letting $M_T$ denote motion between $X$ and $f_T$, $M_T$ can be represented by a transformation sequence $\Phi_T$ containing $p$ transformations. Letting $\odot$ denote the operation of apply transformation sequence to image, we have $ f_T = \Phi_T \odot X$. Letting $\Phi$ represent the group of transformation sequences of all videos, we have $V = \Phi \odot X$. By introducing $G$ implicitly modeling $P_G(\Phi \mid X)$ in transformation space, we have a new description of target:
\begin{equation}
\hat V = G_{\theta}(X) \odot X  \sim P_T(\Phi \odot X \mid X) = P_G(\Phi \mid X) \odot X 
\end{equation}

To make diversity samplings of $V$ feasible, we introduce latent variable $z$ that follows a specific distribution (e.g. Guassian Distribution). Hence, we can modify target of $G_\theta$ from  modeling the distribution $P_G(\Phi \mid X)$ to modeling $P_G(\Phi \mid X, z)$.
This implicit distribution allow us to sample different imaginary videos $\hat V$ through sampling different $z$. Therefore, everything reduces to the following target:
\begin{equation}
\hat V = G_{\theta}(X, z) \odot X \sim P_G(\Phi  \mid X, z)\odot X 
\end{equation}

\subsection{Transformation Generator}
The job of transformation generator is implicitly modeling $P_G(\Phi \mid X, z)$ so that it can generate transformation conditioned on image. Given the condition code of a static image $X$, together with a latent variable $z$, the goal of transformation generator is learning to generate a transformation group $\Phi$.

To be specific, transformation generator outputs $T$ transformation sequences $\{\Phi_1, \Phi_2, \cdots, \Phi_T\}$ corresponding to transformations between $X$ and $\{f_1, f_2, \cdots,  f_T\}$. Each transformation sequence $\Phi_T$ contains $P$ transformations formed by $K$ parameters. Transformations are generated in a sequential fashion in hope of a better description of warp motion, because motion can often be decomposed in a layer-wise manner.

We use two kinds of transformations to model motion. Since in adversarial training, the gradient back-propagation starts from critic network then flows to the frames, the transformation type we choose needs to allow gradient propagating from transformed images to transformation generator. Fortunately, prior works in \cite{jaderberg2015spatial,de2016dynamic} revealed that there is a group of transformations having this adorable attribution. We build two distinct models to form $\Phi$ based on prior works.

\textbf{Affine Transformation: }Simply formed by 6 parameters, affine transformation can model motions including translation, rotation, zoom, and shear. Works in \cite{wang1993layered,bergen1990computing} have shown that affine transformation provides a good approximation of 3-D moving objects motion. Affine transformation works on coordinates, which would raise a problem of undefined pixel locations. In practice, we use differentiable bilinear interpolation to complete those pixels.

\begin{figure}
	\includegraphics[width=3in]{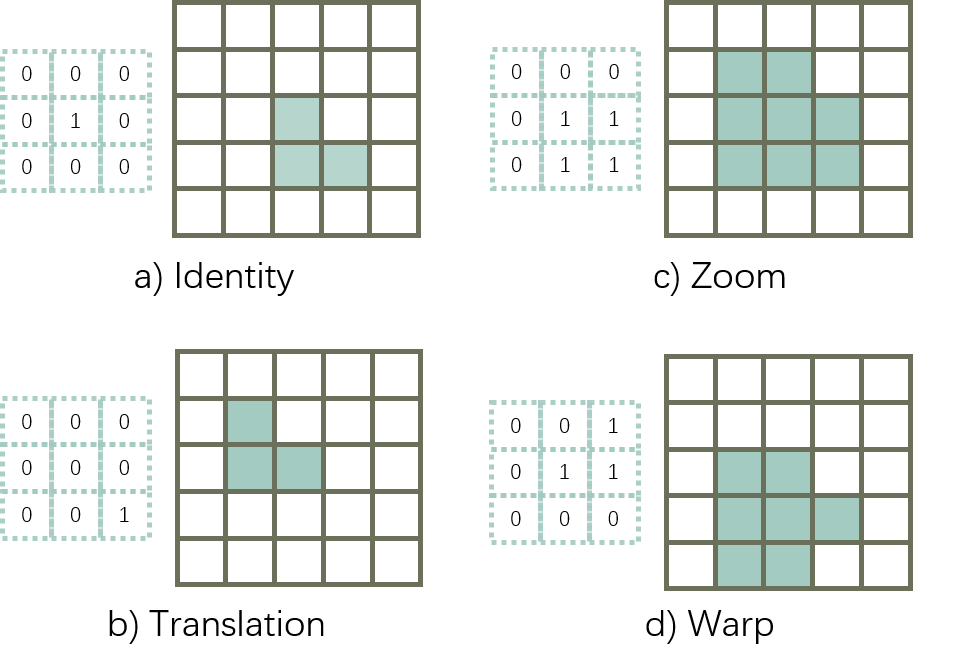}
	\caption{Different convolution kernels result in different motions. \textrm{The doted square denotes convolution kernel and the right side image shows the result of applying the kernel. One simple kernel can model motion like b) translation c) Zoom d) Warp.}}
	\label{fig:filter}
\end{figure}

\textbf{Convolutional Transformation: }A convolution kernel can naturally model simple motions like translation, zoom and warp as shown in Figure \ref{fig:filter}. The kernel size can vary with application scene. For example, a $5 \times 5$ kernel allows pixels translating over a distance of 2 pixels. A sequence of kernel would raise the size of receptive field and allow more complex or intenser motions.


\subsection{Volumetric Merge Network }
Volumetric merge network is responsible for reconstructing frames $\{f_1, f_2, \cdots,  f_T\}$ based on the generated transformation $\Phi$ and image $X$. The transformation group $\Phi$ is finally applied to image $X$, producing an intermediate image group $I$ consisting of $T$ intermediate image sequences $\{{I_1}, {I_2},$ $ \cdots, {I_T}\}$ that will be used to reconstruct $\{f_1, f_2, \cdots,  f_T\}$ accordingly. Combining frames temporally, volumetric merge network outputs imaginary video $\hat V$.

\begin{figure}
\includegraphics[width=3in]{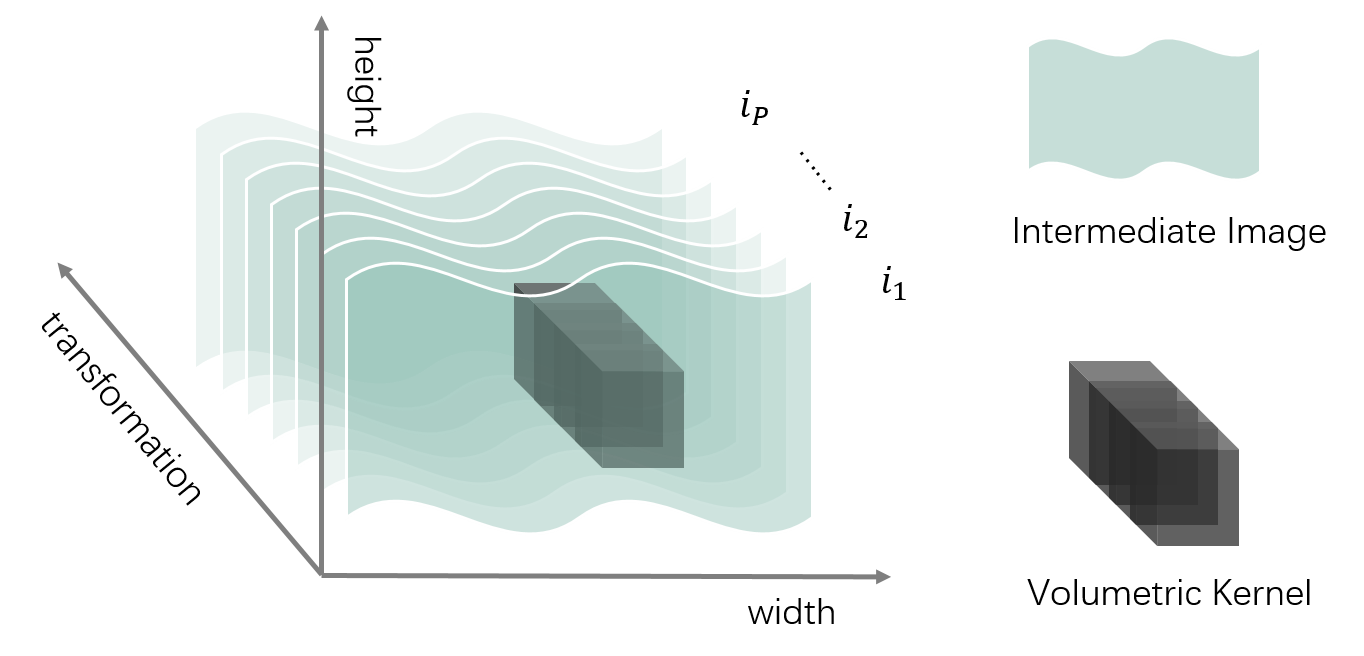}
\caption{Intermediate image sequence $I_T$ as 3-D entity. \textrm{A volumetric kernel can take both neighbor pixel values and intermediate image differences into consideration.}}
\label{fig:volum}
\end{figure}

Since the transformation is generated in a sequential fashion, it is intuitive to take the sequence of intermediate images as an extended dimension representing transformation.
That is, we consider each transformed sequence ${I_T}$ as one entity ${I_T} \in \mathbb{R}^{W \times H \times P}$ that has 3 dimensions as width $W$, height $H$, and transformations $P$.
This 3-D entity, as shown in Figure \ref{fig:volum}, allows us to reconstruct frame by merging it in a volumetric way. Each pixel is reconstructed through volumetric kernels. The kernels can take both neighbor pixel values and intermediate image differences into consideration. 

Parameters in volumetric kernels can be obtained either from clipping a crop of generated transformations or through a specific volumetric kernel generator as shown in Figure \ref{fig:model}. Volumetric kernel generator (a full convolution network) concentrates more on capturing the dependency in spatial domain, while generated transformations can give volumetric kernel better understanding of correlation  between intermediate images. 

\subsection{Video Critic Network} 
To meet the requirement of a better criterion, we design a video critic network $Critic$ to achieve adversarial training. Video critic network $Critic$ receives synthesized video $\hat V$ and real video $V$ as input alternatively, and outputs criticism judging how convincing the input is.  

A convincing video means that the frame looks clear and the motion between frames seems consecutive and reasonable. 
Video critic network $Critic$ needs to give reference of whether the input is plausible and realistic, which requires understanding of both static appearance and dynamic scene. The similar requirement can be found in action recognition task, where lately researchers have made progress \cite{tran2015learning}. We draw inspiration from those works, and design $Critic$ to have the structure of  spatial-temporal convolution networks \cite{ji20133d}.

\begin{figure*}[h]
	\includegraphics[width=7in]{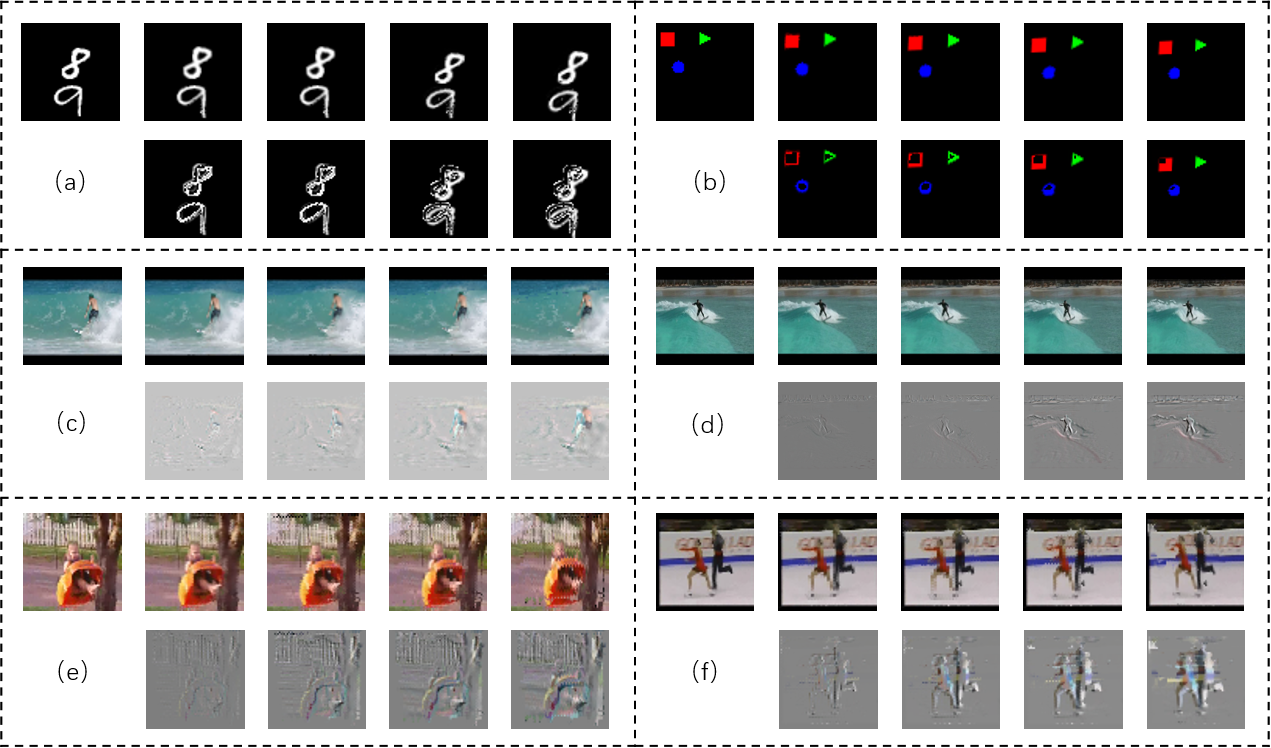}
	\caption{Quality Performance of our framework. \textrm{In each dotted box, the first shows the synthesized imaginary videos given the fist frame as input. The second row shows the difference images of synthesized frames and input. (a)(b) demonstrate the results experiment on moving MNIST and 2D shapes dataset. (c)(d) shows the result of surfing class on UCF101 dataset in different resolution as c) $64 \times 64$ and d) $128 \times 128$. (e)(f) shows the results given image  from swing and ice-dancing categories in UCF101 dataset. The synthesized frames are sharp and clear. Difference images illustrate plausible motions. Results of different resolutions and different image categories on UCF101 dataset suggest our framework shows scale to the complexity of high-resolution videos. }}
	\label{fig:result4in1}
\end{figure*}

\subsection{Learning and Implementations}
Our framework consists of fully feed-forward networks. 
The transformation generator consists of 4 fully connected layers. The latent code sampled from a guassian distribution has 100 dimensions, and the condition code has 512 dimensions. We can encode $X$ into condition code either through refined AlexNet or a 5 layer convolutional network. 
The volumetric merge network consists of 3 volumetric convolutional layers, while the last layer uses element-wise kernel. 
We use a five-layer spatio-temporal convolutional network as the critic network. 

We employ Wasserstein GAN \cite{arjovsky2017wasserstein} to train our framework. The generator loss $L_g$ is defined as: 
\begin{equation}
	L_g = - \mathbb{E}_{v \sim P_T(V \mid X)}  Critic(v)
\end{equation}
The critic loss $L_c$ is defined as: 
\begin{equation}
L_c = \mathbb{E}_{v \sim P_T(V \mid X)}  C(v) - \mathbb{E}_{v \sim P(V \mid X)} Critic(v)
\end{equation}

Alternatively, we minimize the loss $L_g$ once after minimizing the loss $L_d$ 5 times until a fixed number of iterations. 
Ultimately, the optimal video critic network $C$ is hoped to produce good estimate of Earth-Mover (EM) distance between $P(V \mid X)$ and $P_T(V \mid X)$.
We use the RMSProp optimizer and a fixed learning rate of 0.00005. ReLU activation functions and batch normalization are also employed.
%

We use a Tesla K80 GPU and implement the framework in TensorFlow \cite{abadi2016tensorflow}. Our implementation is based on a modified version of \cite{radford2015unsupervised}, and the code can be found at the project page\footnote{https://github.com/gitpub327/VideoImagination} \footnote{ This page contains no information about the authors}. Since we model in relatively small transformation space, the model converges faster than others. Training procedure typically takes only a few days even hours depending on datasets.

\section{Experiment}

In this section, we experiment our framework on 3 video datasets: Moving MNIST \cite{srivastava2015unsupervised}, 2D shape \cite{xue2016visual} and UCF101 \cite{soomro2012ucf101}. For evaluations, we perform qualitative inspection and novel quantitative assessment $RIQA$ to measure the objective quality of the imaginary video.   
\subsection {Baselines and Competing Methods}

Current work about this task is quiet limited. To find out whether our framework outperforms those methods that do not involve our crucial components, we develop two simple but reasonable baselines for this task. For the first one, \textbf{Baseline 1}, the transformation generator and volumetric merge network in our original framework are replaced by a generator network that directly outputs flatten pixels. For the second one, \textbf{Baseline 2}, the whole adversarial training procedure including critic network is removed, and the network is trained minimizing $l_2$ loss function. Those two baselines can also be considered as a form of ablation experiments. 

We also consider several latest works as competing methods as shown in \ref{tab:task}. The task setting is distinct, so it is difficult to find evaluation metrics that can fairly compare all those works together, but we make brave attempt later in Section \ref{sub:eva} to compare our framework against some of those works.


\begin{table}[t]
	\caption{Task setting comparison of related work. \textrm{Multiple output means that the method build a probabilistic model and can sample different results. $\ast$ indicates the methods can also experiment on natural scenes like in UCF101 dataset.}}
	\label{tab:task}
	\begin{tabular}{ccc}
		\toprule
		Model&Input&Output\\
		\midrule
		Ours $\ast$ & image & 5 frames(multiple) \\
		Visual Dynamic \cite{de2016dynamic} & image& 1 frame(multiple)\\
		Scene Dynamic \cite{vondrick2016generating} & image& 32 frames \\
		Dynamic Filter \cite{de2016dynamic} & 4 frames&1 frame \\
		Beyond MSE \cite{mathieu2015deep} $\ast$ & 4 frames&1 frame \\
		Video Sequences \cite{van2017transformation} $\ast$ & 4 frames&4 frames \\

		\bottomrule
	\end{tabular}
\end{table}

\begin{figure}[t]
	\includegraphics[width=3in]{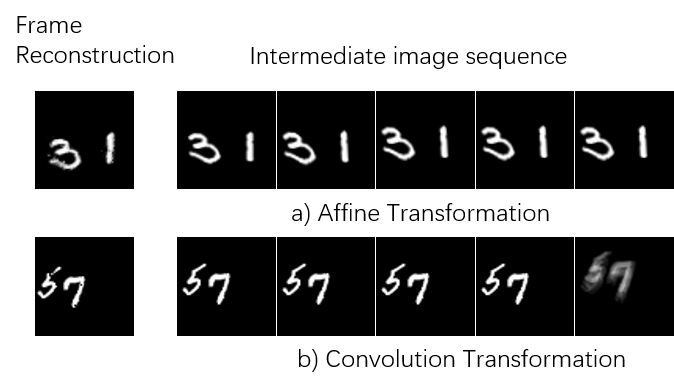}
	\caption{Intermediate image sequences visualization. \textrm{Different transformation models result in different intermediate image sequences $I_T$. Each intermediate image represent one mode of simple transformations. A sequence of these intermediate image can form more complex motion. }}
	\label{fig:intermediate}
\end{figure}
\subsection{Moving MNIST Dataset}
\label{sub:MNIST}
\paragraph{\textbf{Dataset:}}
 We first experiment on a synthetic grey video dataset: moving MNIST dataset \cite{srivastava2015unsupervised}. It consists of videos where two MNIST digits move in random directions with constant speed inside a $64 \times 64$ frame. The 64,000 training video clips and 320 testing clips are generated on-the-fly. Each video clip consists of 5 frames. Taking the first frame as input, the goal is to synthesize multiple imaginary 5-frames videos.

\paragraph{\textbf{Setup:}}
There is barely no pre-processing in our work except for normalizing all videos to be in range $[0, 1]$. We experiment on two transformation models. For convolutional transformation we set kernel size as $9 \times 9$ , and the transformation sequence length $P$ is set as 5 for both models. We generate 4 transformation sequences $\{\Phi_1,\Phi_2,\Phi_3,\Phi_4\}$ corresponding to 4 consecutive frames $\{f_1,f_2,f_3, f_4\}$ at once.

\paragraph{\textbf{Result:}}
Figure \ref{fig:result4in1} (a) illustrates the qualitative performance in moving MNIST dataset. As we can see,  frames are sharp and clear while the shape information of digits is well preserved as we expect. The difference images show that the generated transformations successfully model one motion mode so that the synthesized imaginary video has plausible consecutive motion.
Figure \ref{fig:intermediate} shows reconstructed frames and the corresponding intermediate image sequences in different transformation models.

\subsection{Synthetic 2D Shapes Dataset}
\paragraph{\textbf{Dataset:}}
We experiment our framework using a synthetic RGB video dataset: Synthetic 2D Shapes Dataset \cite{xue2016visual}. There are only three types of objects in this dataset moving horizontally, vertically or diagonally with random velocity in $[0, 5]$. All three objects are simple 2D shapes: circles, squares, and triangles. The original dataset only contains image pairs that have 2 consecutive frames. We extrepolation it to convert image pairs into video clips that have 5 frames. There are 20,000 clips for training and 500 for testing just like settings in \cite{xue2016visual}. We aim at synthesizing multiple imaginary videos each containing five consecutive frames.

\paragraph{\textbf{Setup:}}
The input image size is set as $64 \times 64$ so that we can inherit the network architecture and settings in section \ref{sub:MNIST}. The transformations applied to each color channel are set to be identical for the consistent of RGB channels.

\begin{figure}[t]
	\includegraphics[width=3in]{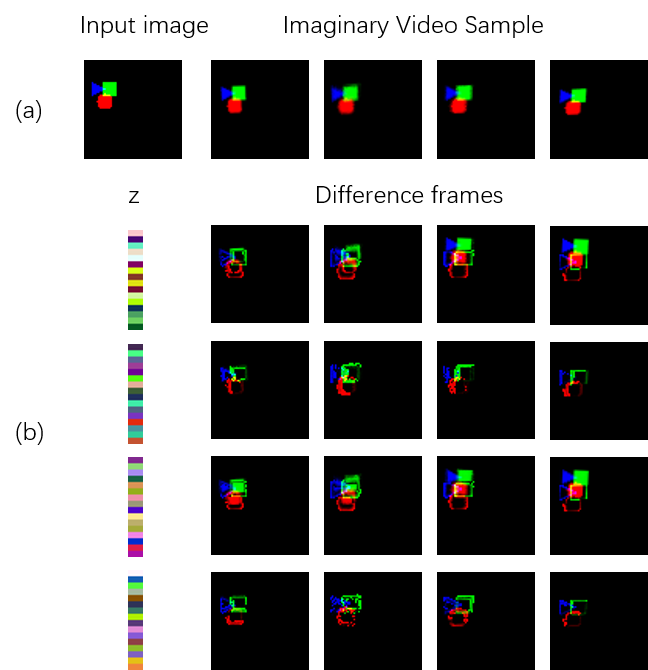}
	\caption{Diverse video imagination: multiple imaginary videos from same input image. \textrm{(a) denotes the input image and one imaginary video sample as reference. The first column of (b) indicates different input $z$s, the rest columns shows the difference frames of imaginary video samples minus the reference. Each row of (b) illustrates a unique imaginary video and its unique $z$. }}
	\label{fig:multi_diff}
\end{figure}

\begin{figure}[t]
	\includegraphics[width=3in]{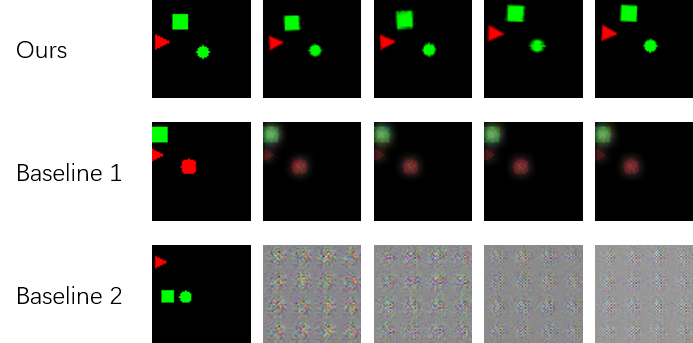}
	\caption{Synthesis result of custom Baselines \textrm{With same fixed training iterations, our framework produce obviously better result. $l_2$ loss in baseline 1 brings blur. Baseline 2 that reconstruct pixel from noise needs much longer training time and cannot produce recognizable frames.}}
	\label{fig:baseline}
\end{figure}

\paragraph{\textbf{Result:}}
Figure \ref{fig:result4in1} (b) illustrates the qualitative performance in 2D shape dataset. Appearance information including color and shape is reconstructed at a satisfying level, and the motion is plausible and non-trivial. 
Multiple sampling results are shown in Figure \ref{fig:multi_diff}. It is clear that sampling different $z$s lead to different imaginary videos with the same input image . Motions in those videos are notably dissimilar. Figure \ref{fig:baseline} gives an perception comparison among our framework and two baselines. The three methods are trained in same iteration. Obviously, generation from scratch as Baseline 2 needs much longer training time and $l_2$ loss criterion as Baseline 1 not only make the result lacking of diversity, but also leads to blur because of intrinsic ambiguity of image. 

\subsection{UCF 101 Dataset}

\paragraph{\textbf{Dataset and setup:}}
The former datasets are both synthetic datasets. For natural scene, we experiment on UCF101 dataset \cite{soomro2012ucf101}. The dataset contains 13,320 videos with an average length of 6.2 seconds belonging to 101 different action categories. The original dataset are labeled for action recognition, but we do not employ those labels and instead we use the dataset in an unsupervised way. Videos with an average length of 6.2 seconds are cut into clips that each consists of five frames. We prepare 15,680 video clips for each category as training samples and 1,000 unseen image as testing samples. The video frames are reshaped to $128 \times 128$ and $64 \times 64$ for different resolution experiments. The convolutional kernel size is set to 16 and 9 accordingly. 


\paragraph{\textbf{Result:}}
Figure \ref{fig:result4in1} (c)(d) illustrate the qualitative performance in surfing class of different resolutions. Obviously our framework produce fairly sharp frames. It successfully escapes from appearance deformation of surfer and wave. The difference images suggest that our framework can model plausible waving and surfing motions. The dynamic results seem rather realistic, so we strongly recommend a quick look at the small gif demo in supplementary material. Figure \ref{fig:wgan} shows the convergence curve of EM distance. We can see the curve decrease with training and converge to a small constant. The absolute of the constant is meaningless because the scale of EM distance varies with architecture of critic network.

\begin{figure}[t]
	\includegraphics[width=3in]{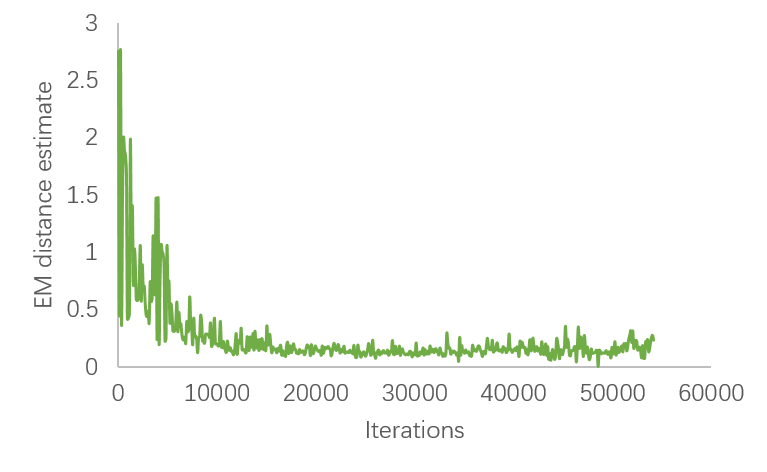}
	\caption{Curve of EM distance estimate at different steps of training. \textrm{The estimation of EM distance is done by a video critic network $C$ that is well trained. We can see that the EM distance decrease and converge with training.} }
	\label{fig:wgan}
\end{figure}

\begin{figure*}
	\includegraphics[width=7in]{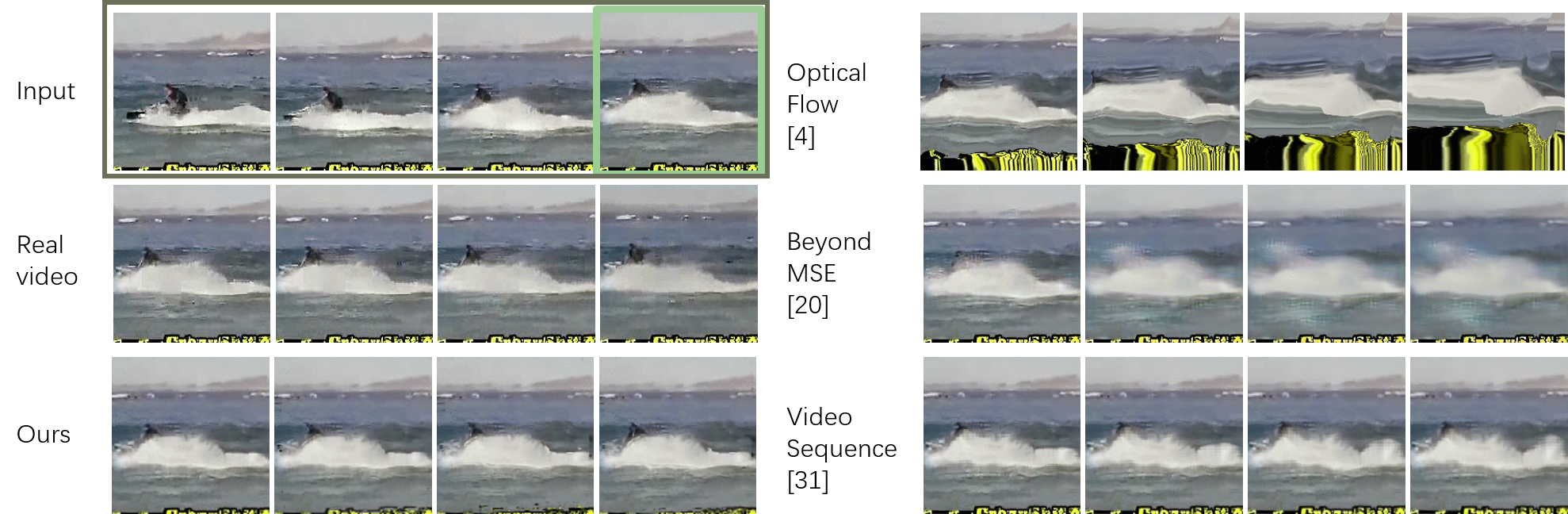}
	\caption{Perceptual Comparison among related works using UCF101 dataset. \textrm{The input frames are from Skijet class. The output frames are reshaped to same size for a fair visual inspection. Notice that our framework only takes one frame as input (the green square) while the rest methods take four frames as input (the grey rectangle). Our result are sharp and relatively clear while the motions of rider and skijet are recognizable and plausible.}}
	\label{fig:methods}
\end{figure*}
\subsection{Evaluation and Comparison:}
\label{sub:eva}

As shown in Table \ref{tab:task}, there is no existing work shares the completely same task settings as ours. To make fair comparison to other works and baseline, we perform both qualitative inspection and novel quantitative evaluation. 

\begin{table}[t]
	\caption{quantitative evaluation comparison among related visual prediction work. \textrm{The lower $RIQA$ indicates better frame reconstruction quality. The BRISQUE score obviously varies with scenes and resolutions. $RIQA$ points out the decreasing proportion between input and output, hence successfully reflects the reconstruction quality.}}
	\label{tab:brisque}
	\begin{tabular}{cccc}
		\toprule
		Methods&\tabincell{c}{Input \\ $BRISQUE$ }&\tabincell{c}{Output \\$BRISQUE$}&\tabincell{c}{$RIQA$}\\
		\midrule
		Ours $64 \times 64$& 45.2164 &47.0168&3.98\% \\
		Ours $128 \times 128$& 35.9809 & 36.7120&\textbf{2.03\%} \\
		Baseline 1 & 45.2164 & 50.7681 &12.28\% \\
		Baseline 2 & 45.2164 & 89.2315 &97.34\% \\
		Optical Flow \cite{brox2004high} & 39.3708 & 40.8481 &3.75\% \\
		Beyond MSE \cite{mathieu2015deep} & 46.3219 & 50.0637& 9.24\% \\
		Video Sequences \cite{van2017transformation} & 39.3708&42.8834&8.92\% \\
		\bottomrule
	\end{tabular}
\end{table}

\textbf{Frame quality assessment.}
Quantitative evaluation of generative models is a difficult, unsolved problem \cite{theis2015note}. The video imagination  task is a multi-modality problem. But traditional full reference image quality assessment methods ($FIQA$) requires a precise ground truth image as reference hence they are no longer appropriate. We employ popular Blind Image Quality Assessment($BIQA$) method $BRISQUE$\cite{mittal2012no} as our non-reference quantitative evaluation metric. 

Since $BRISQUE$ is based on natural 
scene statistic, it is not applicable in synthetic image. we implement it on those methods that can synthesize natural scene images in UCF101 dataset \cite{brox2004high,van2017transformation,mathieu2015deep}. A key problem of employing this metric is that the scenes and resolutions of the synthesized videos may be varied, so it is unfair to make comparison among those samples directly. Fortunately, the quality of the input image can be a solid quality reference. We calculate the decreasing proportion of quality score between inputs and outputs, and take it as our assessment metric: Relative image quality assessment ($RIQA$). 
\begin{equation}
RIQA = \frac{BRISQUE(Input)-BRISQUE(Output)}{BRISQUE(Input)}
\end{equation}
It is fair and reasonable because $RIQA$ eliminates the natural quality differences between scenes and resolutions while have the ability of reflecting the crucial reconstruction quality well.

As shown in Table \ref{tab:brisque}, diversity of the scenes and resolutions makes the raw BRISQUE score not comparable, but the $RIQA$ tells the reconstruction quality change. We can see that our framework outperform other methods, and the poor performances of baselines suggest the architecture of our framework is reasonable. In addition, our framework and Video Sequence\cite{van2017transformation}, that are based on transformation space, do produce images with better qualities than \cite{mathieu2015deep}, which reconstruct frames from scratch.

Table \ref{tab:para} shows the results when we change the hyper-parameters and some model settings, including the number of parameters $K$ forming transformation, the sequence length of transformation $P$ for reconstructing one frame, and the type of transformations. The results demonstrate that our framework is overall robust to those choice. It seems that affine transformation model with transformation sequence length $P=5$ can achieve the best performance.

%

\begin{table}[t]
	\caption{Analysis of the settings of models and hyper-parameters. \textrm{$K$ refers to the number of parameters forming transformation. $P$ refers the sequence length of transformation for reconstructing one frame.}}
	\label{tab:para}
	\begin{tabular}{cc}
		\toprule
		Settings& $RIQA$\\
		\midrule
		\tabincell{c}{affine transformation \\ with $K = 6$ and $P=5$}& \textbf{2.03\%}\\
		\tabincell{c}{affine transformation \\ with $K = 6$ and $P=10$}&4.79\%\\
		\tabincell{c}{convolutional transformation \\ with $K = 8 \times 8$ and $P=5$ }&4.03\%\\		
		\tabincell{c}{convolutional transformation \\ with $K = 16 \times 16$ and $P=5$ }&4.01\%\\
		\bottomrule
	\end{tabular}
\end{table}
\textbf{Qualitative inspection.} Figure \ref{fig:methods} shows the perceptual comparison between our framework and three competing methods \cite{brox2004high,van2017transformation,mathieu2015deep} that also experimenting on UCF101 dataset. Our framework produces four frames conditioned on one frame while other methods take a sequence of four frames as inputs. The simple optical method \cite{brox2004high} fails due to the strong assumption of constant flow speed, yet it perform relatively better in quantitative evaluation because the image get weird but still maintain sharp. Beyond MSE \cite{mathieu2015deep} maintains some appearance but still struggles in deformation and blur. The transformation-based model \cite{van2017transformation} provides fairly recognizable result but 
also gets blurry. Considering \cite{van2017transformation} actually takes four frames as input and aims to predict future frames, the motion looks less consecutive and convincing. Our framework synthesizes sharp and recognizable frames, and the dynamic scene looks realistic and plausible. The motion in our result (wave raising) is not identity to motion in the real video (wave falling), this is because the intrinsic ambiguity of one simgle image. 
Notice that the yellow symbols on the bottom turn to pieces in our framework while in \cite{van2017transformation} it remains still. We believe this is because \cite{van2017transformation} splits frame into patches so gains better description of patch variance. 

\textbf{Failure Case.} A typical failure case in affine transformation model is that the motions between frames are plausible yet unexpected black pixels appear somewhere in the frames. We think this is caused by the empty pixels in intermediate images after applying affine transformations. In convolution model, one common failure mode is that some part of the objects lack resolution while the silhouettes remain recognizable. We believe a more powerful merge network would be a promising solution in both cases,and we leave this for future work.
\section{Conclusion}
In this paper, we have presented a new framework to synthesize multiple videos from one single image. Specifically, our framework uses transformation generation to model the motions between frames, and reconstructs frames with those transformations in a volumetric merge network. We also present a novel evaluation metric to assess the reconstruction quality. We have demonstrated that our framework can produce plausible videos with state-of-the-art image quality on different datasets.

\bibliographystyle{ACM-Reference-Format}
\bibliography{sigproc}

\end{document}